\documentclass[conference]{IEEEtran}
\IEEEoverridecommandlockouts

\usepackage{cite}
\usepackage{amsmath,amssymb,amsfonts}
\usepackage{algorithmicx}
\usepackage{algorithm}
\usepackage{algpseudocode}
\usepackage{float}
\usepackage{courier}
\usepackage{graphicx}
\usepackage{textcomp}
\usepackage{xcolor}
\usepackage{framed}
\usepackage[numbers,sort&compress]{natbib}
\usepackage{multicol}
\usepackage{multirow}
\usepackage{booktabs}
\usepackage{hyperref}
\usepackage{balance}
\def\BibTeX{{\rm B\kern-.05em{\sc i\kern-.025em b}\kern-.08em
		T\kern-.1667em\lower.7ex\hbox{E}\kern-.125emX}}

\newcounter{as}
\newcounter{th}
\begin{document}
	
	\title{Sharpness-Aware Minimization with Adaptive Regularization for Training Deep Neural Networks
		\thanks{*Corresponding authors. This work is sponsored in part by the National Natural Science Foundation of China under Grant  62376278, Hunan Provincial Natural Science Foundation of China (No. 2022JJ10065), Young Elite Scientists Sponsorship Program by CAST (No. 2022QNRC001), Continuous Support of PDL (No. WDZC20235250101).}
	}
	
	\author{\IEEEauthorblockN{Jinping Zou, Xiaoge Deng*, and  Tao Sun*}
		\IEEEauthorblockA{\textit{College of Computer Science and Technology},
			\textit{National University of Defense Technology},
			Changsha, China \\
			Emails: {\{zoujinping, dengxg\}@nudt.edu.cn}, suntao.saltfish@outlook.com}
	}
	\maketitle
	
	\begin{abstract}
Sharpness-Aware Minimization (SAM) has proven highly effective in improving model generalization in machine learning tasks. However, SAM employs a fixed hyperparameter associated with the regularization to characterize the sharpness of the model. Despite its success, research on adaptive regularization methods based on SAM remains scarce. In this paper, we propose the SAM with Adaptive Regularization (SAMAR), which introduces a flexible sharpness ratio rule to update the regularization parameter dynamically. We provide theoretical proof of the convergence of SAMAR for functions satisfying the Lipschitz continuity. Additionally, experiments on image recognition tasks using CIFAR-10 and CIFAR-100 demonstrate that SAMAR enhances accuracy and model generalization.
	\end{abstract}
	
	
	\section{Introduction}
 Since the introduction of Sharpness-Aware Minimization (SAM) by \cite{DBLP:conf/iclr/ForetKMN21}, it has garnered significant attention in machine learning. SAM is a novel and effective method for training overparameterized models and improving generalization performance. By minimizing the maximum loss within the neighborhood of model parameter spaces, SAM achieves a smoother and flatter loss surface, leading to enhanced generalization.
SAM can be viewed as a variant of Stochastic Gradient Descent (SGD), where a fixed hyperparameter is used to capture the sharpness property of the model through regularization. This naturally raises the following question: 
\begin{center} \emph{Can we develop a SAM method with adaptive regularization to further improve the generalization?} 
\end{center}
Despite extensive research on SAM and adaptive methods, no existing studies or algorithms have addressed SAM-based adaptive regularisation to the best of our knowledge. To fill this gap, we propose the SAM with Adaptive Regularization (SAMAR) algorithm. This paper answers the posed question by showing that SAMAR can adaptively adjust regularization parameter based on sharpness information from the loss surface.
  \subsection{Contributions}
The main contributions of this paper are as follows: \begin{enumerate} 
\item We propose a novel variant of the SAM algorithm, SAMAR, which incorporates adaptive regularization. SAMAR adjusts the regularization parameter dynamically based on the ratio of sharpness between successive iterations, leading to better generalization performance. 

\item We provide a theoretical convergence analysis of SAMAR, showing that it achieves a sublinear convergence rate of $\mathcal{O}(1/\sqrt{K})$. 

\item Through empirical experiments on the CIFAR-10 and CIFAR-100 datasets, we demonstrate that SAMAR outperforms other optimizers in terms of accuracy and generalization. 
\end{enumerate}

 \subsection{More Related Works}
 Adaptive Sharpness-Aware Minimization (ASAM), introduced by \cite{kwon2021asam}, incorporates normalization operations and adaptive sharpness to demonstrate that adaptive sharpness exhibits scale-invariant properties. It also shows that adaptive sharpness correlates more strongly with generalization compared to sharpness defined in a fixed-radius spherical region. The authors also provide ASAM's generalization bound derived from the PAC-Bayesian framework, which closely aligns with those of SAM. Several variants of SAM and ASAM have since been developed. 
Li et al. \cite{DBLP:conf/nips/LiG23} introduce the Variance-Suppressed Sharpness-Aware Optimization (VaSSO) scheme, which enhances model stability by reducing the mean square error in the stochastic direction. They define a new $\delta$-stability measure to assess the effect of stochastic linearization. 
Liu et al. \cite{Liu_2022_CVPR} propose the LookSAM and Look-LayerSAM algorithms. LookSAM reduces computational overhead by leveraging the fact that $\boldsymbol{\mathit{g_v}}$, the orthogonal component of SAM’s gradient $\boldsymbol{\mathit{g_s}}$ relative to SGD’s gradient $\boldsymbol{\mathit{g_h}}$, changes slowly, thus allowing for periodic computation of SAM’s gradient. Look-LayerSAM scales up the batch size to 64K for large-batch training.
Li et al. \cite{li2024friendly} present FSAM and reveal that stochastic gradient noise plays a role in improving generalization by decomposing the sample gradient. Mi et al. \cite{mi2022make} propose SSAM, a method with sparse perturbations generated from a binary mask based on Fisher information and dynamic sparse training.
Andriushchenko et al. \cite{Andriushchenko2022} offer a comprehensive understanding toward SAM and explain why SAM converges a solution with better generation performance. Jiang et al. \cite{DBLP:conf/iclr/JiangNMKB20} investigate 40 complexity measures and suggest that sharpness-based measures correlate highly with generalization ability. Shirish Keskar et al. \cite{DBLP:conf/iclr/KeskarMNST17} show that large-batch methods converge to sharp minimizers with poorer generalization and small-batch methods converge consistently to flat minimizers with better generalization. They attribute this discrepancy to the inherent noise in the gradient estimation. Khanh et al. \cite{Khanh2024} provide detailed analysis and rigorous proofs of SAM’s convergence for convex and nonconvex objectives.

In addition to these approaches, there are several powerful methods aimed at improving neural network training and reducing generalization error, such as Adaptive Stochastic Gradient Descent (ASGD) \cite{Sun2020,sun2022adaptive}, the Momentum method \cite{Sun2021,Sun2024,sun2019general}, adaptive step-size methods \cite{li2021second}, the Heavy Ball (HB) algorithm \cite{Wen2023}, and various regularization methods \cite{kohler2017sub, Bellavia2022, Agarwal2021}.
For example, Bellavia et al. \cite{Bellavia2022} construct a third-order polynomial model with an adaptive regularization term $\sigma_k$, using inexact gradient estimates and Hessian information. They show that the expected number of iterations to reach an $\epsilon$-approximation of the first-order stationary point is at most $\mathcal{O}(\epsilon^{-3/2})$. Agarwal et al. \cite{Agarwal2021} extend Adaptive Regularization Cubics (ARC) to optimization on Riemannian manifolds, updating $\sigma_k$ with a more flexible rule.
Bottou et al. \cite{Bottou2018} provide a comprehensive overview of optimization theory in machine learning, offering a holistic view of key methods such as stochastic gradient methods, first-order and second-order regularization techniques, and accelerated gradient methods.
	
\section{Methodology}
	\subsection{Sharpness-Aware Minimization with Adaptive Regularization}
	Our focus is on solving the empirical risk minimization problem in machine learning, which could be formally described as the following process
	\begin{align*}
		\min_{\mathbf x\in\mathbb{R}^d}f(\mathbf{x})\triangleq\frac{1}{n}\sum_{i=1}^n f_i(\mathbf{x}),\ \text{where }f_i(\mathbf{x})\triangleq\ell\big ( h(s_i;\mathbf{x}),y_i\big).
	\end{align*}
 
	Taking the sample $s_i$ as input, the prediction function $h(s_i;\mathbf{x})$ parameterized by the model parameter $\mathbf{x}\in \mathbb R^d$ together with the sample label $y_i$ constitute the input of loss function $\ell\big ( h(s_i;\mathbf{x}),y_i\big)$, which represents the loss caused by the model parameter $\mathbf{x}$ at the i-th sample label-pair $(s_i,y_i)$. The total empirical loss $f(\mathbf x)$ is defined as the arithmetic mean of losses of all sample label-pairs, $\frac{1}{n}\sum_{i=1}^n f_i(\mathbf{x})$.
	
    Traditionally, the optimization procedure aims to attain the most suitable model parameter $w_*$ through a numerical iterative algorithm that minimizes the empirical loss on the training set. However, in various cases, such a model parameter $w_*$ does not yield a satisfactory test accuracy on the test dataset while leading the loss landscape with sharp local minima and leaving the model with poor generalization performance.
	
    Therefore, given narrowing the generalization gap, inspired by the adaptive regularization method in \cite{Agarwal2021,Bellavia2022} and the definition of sharpness in \cite{DBLP:conf/iclr/ForetKMN21}, letting $R(\mathbf{x})= \underset{\| \boldsymbol{\epsilon} \| \leq \rho}{\max} f(\mathbf{x}+\boldsymbol{\epsilon} ) -f ( \mathbf{x})$, the objective function in our proposed SAM with adaptive regularization parameter (SAMAR) is defined as follows
	\begin{align}
		\underset{\mathbf{x}\in\mathbb{R}^d}{\min}f(\mathbf{x}) \!+\! \lambda R(\mathbf{x}) &= \underset{\mathbf{x}\in\mathbb{R}^d}{\min} \underset{\| \boldsymbol{\epsilon} \| \leq \rho}{\max}(1\!-\!\lambda)f(\mathbf{x}) \!+\! \lambda f(\mathbf{x}\!+\!\boldsymbol{\epsilon}), \label{objfunc}
	\end{align}
	where $\rho$ is the radius of the neighborhood near $\mathbf x$. Apparently $\underset{\mathbf{x}\in\mathbb{R}^d}{\min}f(\mathbf{x}) +\lambda R(\mathbf{x})$ represents a trade-off between minimization loss and sharpness reduction. We not only emphasize reducing the training loss as much as possible but also consider strengthening the model generalization performance. The regularization parameter $\lambda$ reflects the importance placed on the generalization capability.
    
	Next, consider the internal maximization procedure of \eqref{objfunc}. Let $\boldsymbol{\epsilon}_{\mathbf{x}_{k}}$ represent the maximal point of the internal maximization problem at k-th step $\mathbf{x}_k$. Then, the following consequences could be derived 
	\begin{align*}
		\boldsymbol{\epsilon}_{\mathbf{x}_{k}} &\triangleq \underset{\| \boldsymbol{\epsilon} \| \leq \rho}{\arg \max}\;(1 -\lambda_{k})f(\mathbf{x}_{k})+\lambda_{k}f(\mathbf{x}_{k}+\boldsymbol{\epsilon}) \\
		&\stackrel{(\star)}{\approx} \underset{\| \boldsymbol{\epsilon} \| \leq \rho}{\arg \max} \;\lambda_{k} \left [f(\mathbf{x}_{k})+\nabla f(\mathbf{x}_{k})^T\boldsymbol{\epsilon} \right]\\ 
		&\stackrel{(\triangle)}{\approx} \underset{\| \boldsymbol{\epsilon} \| \leq \rho}{\arg \max} \;\lambda_{k} \left [f(\mathbf{x}_{k})+ \mathbf{g}(\mathbf{x}_{k})^T\boldsymbol{\epsilon} \right], \\
		&\Rightarrow \boldsymbol{\epsilon}_{\mathbf{x}_{k}} = \rho \frac{\mathbf{g}(\mathbf{x}_k)}{\| \mathbf{g}(\mathbf{x}_k) \|} ,
	\end{align*}
	where $(\star)$ is first-order approximated via expanding $f(\mathbf{x}_{k}+\boldsymbol{\epsilon})$ at $\mathbf{x}_{k}$ with Taylor's formula. $(\triangle)$ is approximated by utilizing the stochastic gradient of the sample $\mathbf{g}(\mathbf{x}_k)$ instead of the full gradient $\nabla f(\mathbf{x}_{k})$. This result of adversary perturbation aligns with \cite{mi2022make, li2024friendly, Liu_2022_CVPR}. Thus, referring to the optimization process of SGD and SAM,  our object function \eqref{objfunc} is equivalent to solving the following minimization problem \eqref{new_obj} $(\boldsymbol{\epsilon}_{\mathbf{x}}=\rho \frac{\mathbf{g}(\mathbf{x})}{\| \mathbf{g}(\mathbf{x}) \|}, \mathbf{x}\in\{\mathbf x_0, \mathbf x_1,...,\mathbf x_{K-1} \})$. Then, the process of the SAMAR algorithm can be described as Algorithm \ref{alg.1}. 
	\begin{align} \label{new_obj}
		f(\mathbf{x}) + \lambda_k R(\mathbf{x})\approx(1-\lambda_k)f(\mathbf{x}) + \lambda_k f(\mathbf{x}+\boldsymbol{\epsilon}_{\mathbf x}).
	\end{align}

	The initialization of hyperparameters $(\lambda_0, \chi, \gamma,...,)$, the random sampling and adversary perturbation calculation in steps 1-2 are ordinary. The parameter of the model $\mathbf{x}_{k+1}$ is updated iteratively utilizing gradient descent in steps 3-4. Steps 5-6 capture the adaptive update rule for the regularization parameter $\lambda_{k+1}$. In Algorithm \ref{alg.1}, $\chi$ represents the threshold for the ratio of the recent sharpness to the previous one. $\gamma$ restricts the amplitude of the adaptive adjustment to $\lambda_{k+1}$. $\textbf{Proj}_{[\delta,1-\delta]}(\cdot)\triangleq \max\big(\delta,\min(\cdot,1-\delta)\big)$ denotes a projection operation, which constrains the range of $\lambda_{k+1}$ to $[\delta,1-\delta]$.

 \begin{algorithm} 
		\caption{Sharpness Aware Minimization with Adaptive Regularization} \label{alg.1}
		\begin{algorithmic}
			\State \textbf{Initialize: }$\mathbf{x}_{0},\rho,\eta_0,0<\lambda_{0}<1,\chi,\gamma>1,0<\delta<1$.
			\For{$k = 0,\dots,K-1$} 
			\State\textbf{step 1: }Compute the stochastic gradient $\mathbf{g}(\mathbf{x}_{k})$ on the \\
                \qquad \qquad \quad batch data.
			\State\textbf{step 2: }Calculate $\boldsymbol{\epsilon}_{\mathbf{x}_{k}}=\rho \dfrac{\mathbf{g}(\mathbf{x}_k)}{\| \mathbf{g}(\mathbf{x}_k) \|} $ through first-order\\
			\qquad \qquad \quad stochastic linearization approximation.
			\State \textbf{step 3: }Let  $\mathbf{s}_k=(1-\lambda_{k})\mathbf{g}(\mathbf{x}_{k})+\lambda_{k}\mathbf{g}(\mathbf{x}_{k}+\boldsymbol{\epsilon}_{\mathbf{x}_{k}})$.
			\State \textbf{step 4: } $\mathbf{x}_{k+1}=\mathbf{x}_{k}-\eta_{k}\mathbf{s}_k$.
			\State \textbf{step 5: }Calculate $r_k=\dfrac{\|\mathbf{g}(\mathbf{x}_{k})\|}{\|\mathbf{g}(\mathbf{x}_{k-1})\|}$.
			\State \textbf{step 6: }
			\State \qquad \quad\textbf{if }{$r_k\geq \chi$} or $k=0$\ : $\lambda_{k+1}=\textbf{Proj}_{[\delta,1-\delta]}\big(\gamma\lambda_k \big)$. \\
			\State\qquad\quad\textbf{else}\ :
			$\lambda_{k+1}=\textbf{Proj}_{[\delta,1-\delta]}\big(\frac{1}{\gamma}\lambda_k \big)$.
			\State \textbf{step 7: }$k=k+1$.
			\EndFor	\\
			\textbf{return } $\mathbf{x}_{K}$
		\end{algorithmic}
\end{algorithm}

    \textbf{Adaptive Regularization Parameter Strategy.}
    Our work builds upon previous adaptive regularization strategies introduced by \cite{Agarwal2021,Bellavia2022}, with a focus on dynamically adjusting the regularization parameter $\lambda$ based on sharpness changes during the optimization procedure. The core of our strategy is centred around the following ratio:
\begin{align*}
    r_k & \triangleq \dfrac{|f(\mathbf{x}_{k} + \boldsymbol{\epsilon}_{\mathbf{x}_{k}}) - f(\mathbf{x}_{k})|}{|f(\mathbf{x}_{k-1} + \boldsymbol{\epsilon}_{\mathbf{x}_{k-1}}) - f(\mathbf{x}_{k-1})|}   \\
    & \approx \dfrac{|\langle\mathbf{g}(\mathbf{x}_{k}),\boldsymbol{\epsilon}_{\mathbf{x}_{k}}\rangle|}{|\langle\mathbf{g}(\mathbf{x}_{k-1}),\boldsymbol{\epsilon}_{\mathbf{x}_{k-1}}\rangle|}=\dfrac{\|\mathbf{g}(\mathbf{x}_{k})\|}{\|\mathbf{g}(\mathbf{x}_{k-1})\|},
\end{align*}
which serves as a measure of how much the local sharpness changes. If $r_k$ becomes significantly larger than 1, this indicates that the optimization process has moved into a sharper region of the loss surface. In such cases, increasing the regularization parameter $\lambda$ is necessary to counteract the sharpness and encourage the algorithm to find flatter minima, which generally leads to better generalization performance.
On the other hand, if $r_k$ becomes much smaller than 1, the algorithm has entered a flatter region. In this scenario, reducing $\lambda$ allows the algorithm to focus more on minimizing the primary objective function. This ensures that the model can fit the sample data more effectively without being overly constrained by the regularization term.
	
	\subsection{Convergence Analysis}
	In this subsection, the convergence analysis process of Algorithm \ref{alg.1} will be performed to deduce its rate of convergence. Before we start, the following assumptions are outlined, which are pretty common in the general algorithm convergence analysis process. 

        \refstepcounter{as}
	\textbf{Assumption \arabic{as} (Lower Bound.)}\label{hp1}
		$f(\mathbf{x})$ has a lower bound $f_{\mathrm{inf}}$, i.e., $f(\mathbf{x})\geq f_{\mathrm{inf}} >-\infty\; \forall \ \mathbf{x} \in \mathbb{R}^d$, which means that the optimal value of $f(\mathbf{x})$ exists.

	\refstepcounter{as}
	\textbf{Assumption \arabic{as} (Lipschitz Gradient.)}\label{hp2}
		$f(\mathbf{x})$ is continuously differentiable and $\nabla f(\mathbf{x})$ satisfies the Lipschitz condition. Namely, there exists a positive constant $L>0$ and
$	\|\nabla f(\mathbf x)- \nabla f(\mathbf y)\| \leq L\| \mathbf x -\mathbf y \|, 
			\|\mathbf g(\mathbf{x})-\mathbf g(\mathbf{y})\| \leq L\|\mathbf{x}-\mathbf{y}\|, \ \forall\ \mathbf{x},\mathbf{y} \in \mathbb{R}^d.
$		
	
	\refstepcounter{as}
	\textbf{Assumption \arabic{as} (Bounded Variance.)}\label{hp3}
		The stochastic gradient of the sample $\mathbf{g}(\mathbf{x}_k)$ is the unbiased estimation of $\nabla f(\mathbf x_k)$ and the variance of $\mathbf{g}(\mathbf{x}_k)$ exists, i.e., $
\mathbb{E}[\mathbf{g}(\mathbf{x}_k)]=\nabla f(\mathbf x_k), 
\mathbb{E} \left [ \|\mathbf{g}(\mathbf{x}_k)-\nabla f(\mathbf x_k)\|^2\right ]\leq \sigma^2, 
$		
  for some constant $\sigma>0$.
  
\refstepcounter{th}
\textbf{Theorem \arabic{th} (Convergence Results of SAMAR.)}\label{th1}
	Under the Assumptions \ref{hp1}-\ref{hp3}, $\rho$ and $\nu$ satisfy $\rho=\frac{\rho_0}{\sqrt K}$, $\nu=1-\frac{5L\eta_0}{2\sqrt K}$, for the sequence $\{\mathbf x_k\}_{k\geq 0}$ generated by Algorithm \ref{alg.1} run with a learning rate $\eta_k=\frac{\eta_0}{\sqrt K}\leq \frac{2}{5L}$. Then it holds that
		\setlength{\jot}{6pt} 
		\begin{align*}
			&\dfrac{1}{K}\sum_{k=0}^{K-1}\mathbb E[\|\nabla f(\mathbf x_k)\|^2]\\
			&\leq \frac{1}{\nu}\left [\frac{f(\mathbf x_0)-f_{\mathrm{inf}}}{\eta_0 \sqrt K}+\frac{L\rho_0^2}{2\eta_0\sqrt K}+\frac{3L\eta_0\sigma^2}{\sqrt K} \right .
			\left .+\frac{2L^3\eta_0\rho_0^2}{K^\frac{3}{2}} \right ],
		\end{align*}
  and
		\begin{align*}
			\dfrac{1}{K}\sum_{k=0}^{K-1}\mathbb E[\|\nabla f(\mathbf x_k\!+\!\boldsymbol{\epsilon}_{\mathbf{x}_{k}})\|^2] &\!\leq\! \frac{2}{\nu}\left [\frac{f(\mathbf x_0)-f_{\mathrm{inf}}}{\eta_0\sqrt{K}}+\frac{L\rho_0^2}{2\eta_0\sqrt K} \right .\\
			&\!+\! \left . \frac{3L\eta_0\sigma^2}{\sqrt K}\!+\!\frac{2L^3\eta_0\rho_0^2}{K^\frac{3}{2}} \right ]\!+\!\frac{2L^2\rho_0^2}{K}.
		\end{align*}

        Theorem \ref{th1} indicates that the SAMAR algorithm exhibits a sublinear convergence rate $\mathcal{O}(\frac{1}{\sqrt{K}})$, which is consistent with SGD. The detailed proof of Theorem \ref{th1} can be found in the Appendix\footnote{\href{https://github.com/JinpingZou/SAMAR/tree/main}{See https://github.com/JinpingZou/SAMAR/tree/main.}}. Notably, our proof does not require the more restrictive assumption of bounded gradient compared to the literature \cite{mi2022make, gitman2019understanding, li2024friendly, Sun2024}. 

	\begin{figure}[t]
		\begin{tabular}{cc}
			\includegraphics[width=0.23\textwidth]{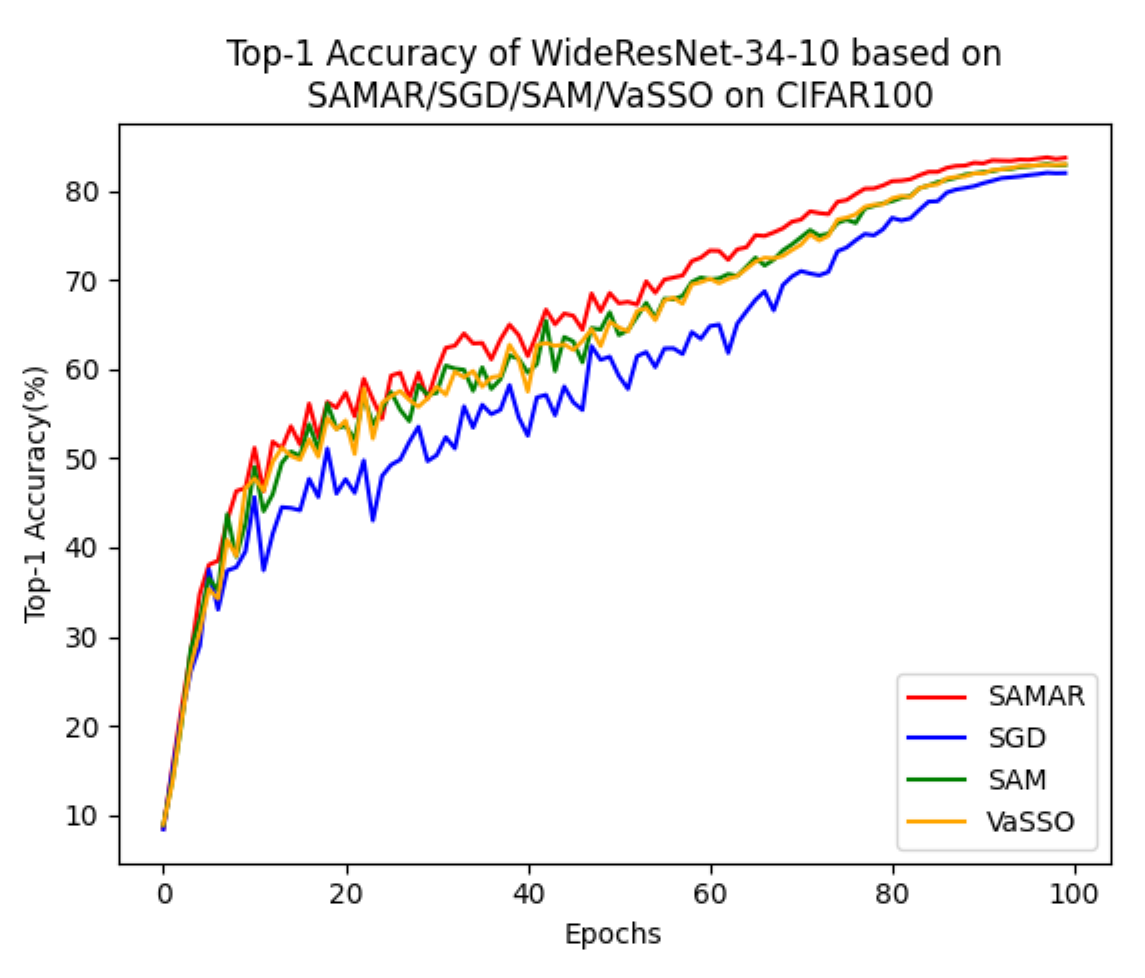}&
			\includegraphics[width=0.22\textwidth]{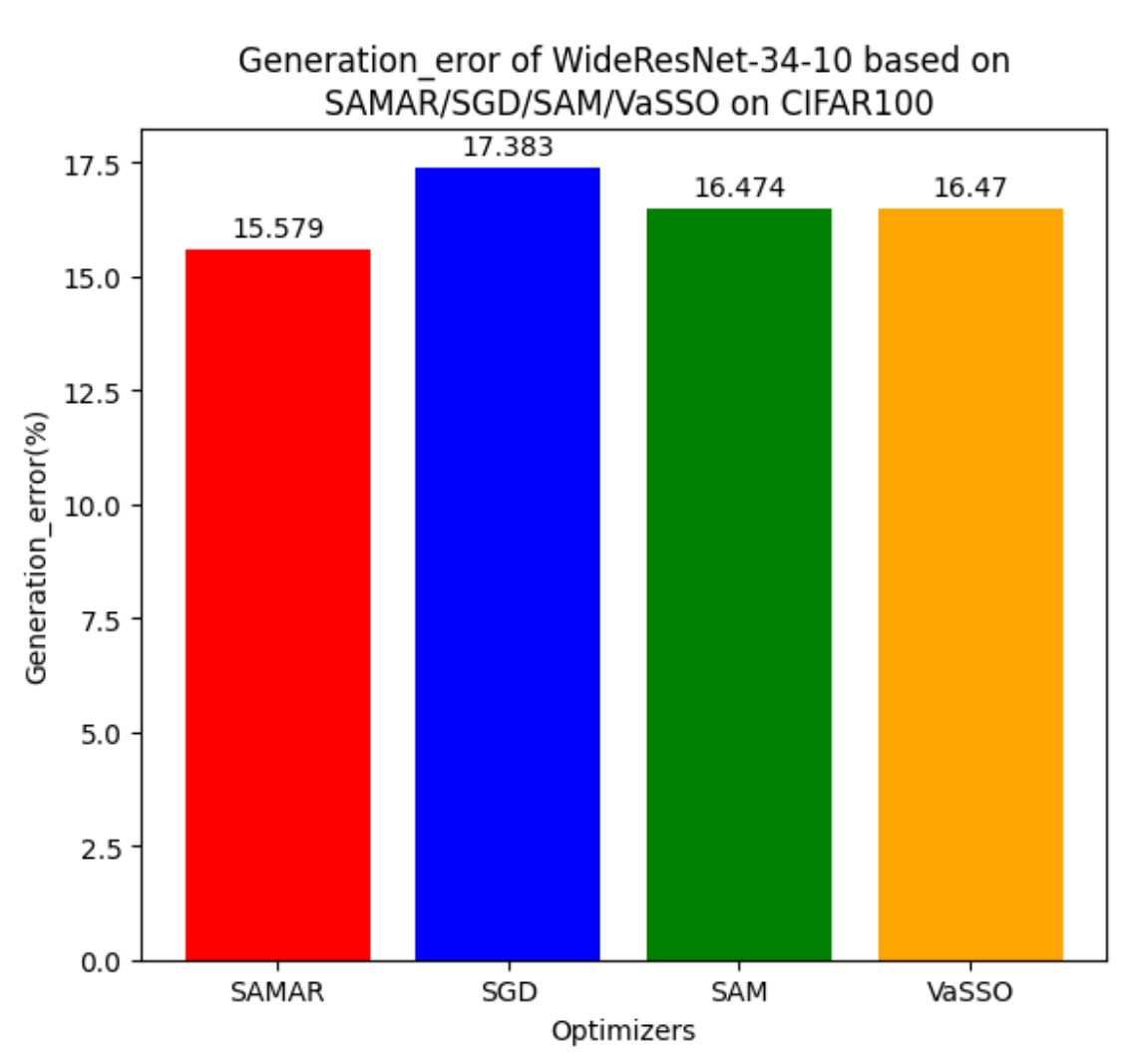}
			\\ 
			(a)  & ~~~~(b) 
			\\
		\end{tabular}
		\caption{(a) Epochs vs. Top-1 accuracy obtained from WideResnet-34-10 for CIFAR-100 classification by utilizing different optimizers; (b) Generation\_error of WideResNet-34-10 based on different optimizers on CIFAR-100.}
		\label{fig.1}
	\end{figure}
	
	\begin{table*}
		\centering
		\caption{Comparison of 5 Evaluation Metrics (\%) for ResNet-34 and WideResNet-34-10 using SAMAR, SGD, SAM, and VaSSO on CIFAR-10}
		\label{cifar10}
		\begin{tabular}{c|c|c|c|c|c}
			\hline
			CIFAR-10  & 		Metrics 	  	 & 				SAMAR 			   	  & 			SGD 		  & 			SAM 	   	  		   & 		VaSSO\\
			\hline
			&  		Top-1 		 & $96.370_{96.297\pm0.054}$ 		  & $95.730_{95.640\pm0.067}$ & $96.190_{96.147\pm0.037}$ 		   & $\textbf{96.500}_{96.320\pm0.131}$\\
			&  		Top-5 		 & $99.950_{99.943\pm0.005}$ 		  & $99.950_{99.933\pm0.017}$ & $99.960_{99.953\pm0.005}$ 		   & $\textbf{99.990}_{99.973\pm0.012}$\\
			ResNet-34 		 &  Last10\_Top-1\_Test  & $\textbf{96.184}_{96.117\pm0.058}$ & $95.414_{95.328\pm0.062}$ & $96.059_{96.020\pm0.028}$ 		   & $96.037_{96.179\pm0.103}$\\
			&  Last10\_Top-1\_Train & $98.802_{98.742\pm0.043}$ 		  & $98.551_{98.504\pm0.054}$ & $98.840_{98.829\pm0.008}$ 		   & $98.820_{98.759\pm0.085}$\\
			& Generation\_Error 	 &	$2.665_{2.625\pm0.031}$ 		  & $3.234_{3.177\pm0.041}$   & $2.839_{2.809\pm0.031}$   		   & $\textbf{2.646}_{2.580\pm0.055}$\\
			\hline
			& 		 Top-1 			 & $\textbf{97.010}_{96.963\pm0.034}$ & $96.600_{96.530\pm0.051}$ & $96.890_{96.837\pm0.075}$ 		   & $96.940_{96.913\pm0.021}$\\
			&  	 Top-5 			 & $\textbf{99.980}_{99.973\pm0.005}$ & $99.960_{99.953\pm0.005}$ & $\textbf{99.980}_{99.973\pm0.009}$ & $\textbf{99.980}_{99.977\pm0.005}$\\
			WideResNet-34-10 &  Last10\_Top-1\_Test  & $\textbf{96.947}_{96.874\pm0.053}$ & $96.524_{96.444\pm0.057}$ & $96.839_{96.777\pm0.071}$ 		   & $96.863_{96.834\pm0.025}$\\
			&  Last10\_Top-1\_Train & $99.420_{99.407\pm0.011}$ 		  & $99.400_{99.388\pm0.009}$ & $99.420_{99.417\pm0.004}$ 		   & $99.424_{99.417\pm0.006}$\\
			& Generation\_Error 	 & $\textbf{2.572}_{2.534\pm0.051}$   & $2.987_{2.944\pm0.057}$   & $2.734_{2.640\pm0.067}$ 		   & $2.615_{2.583\pm0.028}$\\
			\hline
		\end{tabular}
	\end{table*}

	\begin{table*}
		\centering
		\caption{Comparison of 5 Evaluation Metrics (\%) for ResNet-34 and WideResNet-34-10 using SAMAR, SGD, SAM, and VaSSO on CIFAR-100}
		\label{cifar100}
		\begin{tabular}{c|c|c|c|c|c}
			\hline
			CIFAR-100 & Metrics 			  	 & SAMAR 							  & SGD 		  		  	  & SAM 					  & VaSSO\\
			\hline
			&  Top-1 				 & $\textbf{79.980}_{79.633\pm0.256}$ & $78.070_{77.707\pm0.507}$ & $79.220_{79.027\pm0.148}$ & $79.690_{79.350\pm0.453}$\\
			&  Top-5 				 & $\textbf{95.510}_{95.347\pm0.117}$ & $94.650_{94.490\pm0.199}$ & $95.380_{95.253\pm0.111}$ & $95.440_{95.253\pm0.257}$\\
			ResNet-34 		 &  Last10\_Top-1\_Test  & $\textbf{79.564}_{79.270\pm0.235}$ & $77.841_{77.401\pm0.509}$ & $78.912_{78.763\pm0.141}$ & $79.285_{78.999\pm0.398}$\\
			&  Last10\_Top-1\_Train & $98.281_{98.221\pm0.067}$ 		  & $97.669_{97.613\pm0.040}$ & $98.456_{98.412\pm0.040}$ & $98.467_{98.419\pm0.041}$\\
			& Generation\_Error 	 & $\textbf{19.265}_{18.951\pm0.291}$ & $20.981_{20.211\pm0.549}$ & $19.785_{19.649\pm0.101}$ & $19.931_{19.420\pm0.361}$\\
			\hline
			
			&  Top-1 				 & $\textbf{83.770}_{83.753\pm0.024}$ & $82.360_{82.023\pm0.248}$ & $83.010_{82.983\pm0.021}$ & $83.020_{82.980\pm0.029}$\\
			&  Top-5 				 & $\textbf{96.970}_{96.840\pm0.092}$ & $96.070_{95.930\pm0.104}$ & $96.590_{96.487\pm0.101}$ & $96.600_{96.527\pm0.057}$\\
			WideResNet-34-10 &  Last10\_Top-1\_Test  & $\textbf{83.490}_{83.468\pm0.025}$ & $81.911_{81.599\pm0.223}$ & $82.631_{82.605\pm0.031}$ & $82.722_{82.631\pm0.069}$\\
			&  Last10\_Top-1\_Train & $99.020_{99.012\pm0.006}$ 		  & $98.791_{98.780\pm0.010}$ & $99.035_{99.027\pm0.008}$ & $99.040_{99.025\pm0.012}$\\
			& Generation\_Error 	 & $\textbf{15.579}_{15.545\pm0.026}$ & $17.383_{17.180\pm0.217}$ & $16.474_{16.421\pm0.037}$ & $16.470_{16.395\pm0.077}$\\
			\hline

		\end{tabular}
	\end{table*}
	
	\section{Numerical Experiments}
	In this section, in order to verify that SAMAR performs well in improving the model generalization, we conduct a series of experiments to compare SAMAR with SGD, SAM and VaSSO optimizers in terms of effectiveness and model generalization. The code can be found at the same link as the Appendix.
	\subsection{Models, Benchmark Datasets and Metrics}
	CIFAR-10 and CIFAR-100 are widely recognized benchmarks, allowing for fair comparison with existing methods. Hence we train ResNet-34 \cite{He_2016_CVPR} and WideResNet-34-10 \cite{Zagoruyko2016} neural network models on CIFAR-10 and CIFAR-100 datasets. Meanwhile, five evaluation metrics are recorded: the maximum of top-1 accuracy and the maximum of top-5 accuracy on the test dataset during 100 epochs, the average of top-1 accuracy for the last ten epochs on the test and train datasets, and generalization error. They are denoted as Top-1, Top-5, Last10\_Top-1\_Test, Last10\_Top-1\_Train, and Generation\_Error in order of precedence, respectively, where Generation\_Error is computed by differencing Last10\_Top-1\_Train with Last10\_Top-1\_Test. The experimental data in all tables are obtained after three independent runs. The best values for the four metrics related to the test dataset and generalization performance are bolded, and the data format is $\text{max}_{\text{mean}\pm \text{std}}$.
	\subsection{Implementation}
	We train ResNet-34 and WideResNet34-10 for 100 epochs on CIFAR-10 and CIFAR-100 with a batch size of 256. To strengthen the model's capacity for feature learning, we perform data augmentation through normalization, cutout, random cropping and random horizontal flip. To accelerate the speed of convergence of the training process, a learning rate scheduling strategy is adopted during the training process. This scheduling strategy is specified by decaying the learning rate with a CosineAnneal strategy between different epochs and maintaining the same learning rate between different batches of the same epoch. For additional hyperparameters setting, please refer to Table \ref{hyperparameters} in the Appendix for detail. All experiments are performed on a single NVIDIA RTX 4090D GPU.

	\subsection{Analysis of Results}
	\textbf{CIFAR-10.}
	As in the case of the data shown in Table \ref{cifar10}, SAMAR outperforms SAM and SGD in almost all the best values for the four metrics relative to the test data set and generalization performance, provided that the models used are identical. In comparison, the main reason that SAMAR does not perform as well as VaSSO on some metrics is that the scale of CIFAR-10 dataset is relatively small, therefore the models are easily overfitted and do not tend to apparently reflect the performance gap between different optimizers.
	
	
    \textbf{CIFAR-100.}
    Based on the results of the experiment in Table \ref{cifar100}, we can conclude that SAMAR outperforms SAM and VaSSO, especially compared to the SGD optimizer when training the ResNet-34 and WideResNet-34-10 models. Specifically, SAMAR achieves a significant improvement in classification accuracy. In addition, it shows better control of the generalization error, which indicates that SAMAR strengthens the generalization ability of the models while improving their performance. The Top-1 accuracy of SAMAR is about 0.7\% higher than SAM and VaSSO, and the Generalization\_Error is about 0.9\% lower than SAM and VaSSO. The variation of Top-1 accuracy with iterative epochs when training WideResNet-34-10 on CIFAR-100 using different optimizers separately is shown in Fig. \ref{fig.1} (a), while the Generalization\_Error is shown in Fig. \ref{fig.1} (b). Both subfigures illustrate that the usage of SAMAR can produce higher performance gains and stronger generalization ability when training slightly large models on relatively complex datasets.
	
    \section{conclusions}
This paper presents SAMAR, a novel algorithm featuring adaptively adjustable the regularization parameter $\lambda$. Although the convergence rate of SAMAR is theoretically aligned with that of SAM, empirical results on CIFAR-10 and CIFAR-100 datasets demonstrate that SAMAR consistently achieves higher accuracy and superior generalization, especially for larger models on complex datasets. However, we acknowledge that SAMAR may lag behind VaSSO in certain scenarios. Our future work will focus on enhancing the updating mechanisms for adversarial perturbations to further improve generalization performance.

\newpage
\bibliographystyle{plain}

\begin{onecolumn}
\newpage
    \section*{\large\bfseries Appendix for}
    \begin{center}
        \textbf{Sharpness-Aware Minimization with Adaptive Regularization for Training Deep Neural Networks}
    \end{center}
	\subsection{Proof of Theorem 1}

	A significant conclusion \eqref{ineq.1} could be deduced from Assumption \ref{hp2}. If $\nabla f(\mathbf{x})$ has a Lipschitz gradient with constant $L>0$, we have
	\begin{align}\label{ineq.1}
		f(\mathbf{x}_{k+1})-f(\mathbf{x}_{k}) &\leq \langle \nabla f(\mathbf{x}_{k}),\mathbf{x}_{k+1}-\mathbf{x}_{k} \rangle + \frac{L}{2} \Vert \mathbf{x}_{k+1} - \mathbf{x}_{k} \Vert^2   \\
	&=-\eta_{k}\big\langle \nabla f(\mathbf{x}_{k}),(1-\lambda_{k}) \mathbf{g}(\mathbf{x}_{k}) + \lambda_{k} \mathbf{g}(\mathbf{x}_{k}+\boldsymbol{\epsilon}_{\mathbf{x}_{k}}) \big\rangle +\frac{L}{2}\eta_{k}^{2} \big\Vert (1-\lambda_{k})\mathbf{g}(\mathbf{x}_k) + \lambda_{k} \mathbf{g}(\mathbf{x}_{k}+\boldsymbol{\epsilon}_{\mathbf{x}_{k}}) \big\Vert^2    \nonumber     \\
	&=\underbrace{-\eta_{k} \big\langle \nabla f(\mathbf{x}_{k}),(1-\lambda_{k}) \mathbf{g}(\mathbf{x}_{k}) \big\rangle}_{:=\spadesuit} \!-\! \underbrace{\eta_{k}\big\langle \nabla f(\mathbf{x}_{k}),\lambda_{k} \mathbf{g}(\mathbf{x}_{k}+\boldsymbol{\epsilon}_{\mathbf{x}_{k}})\big\rangle}_{:=\heartsuit}\!+\!\underbrace{\frac{L}{2}\eta_{k}^{2} \big\Vert (1-\lambda_{k}) \mathbf{g}(\mathbf{x}_{k}) \!+\! \lambda_{k} \mathbf{g}(\mathbf{x}_{k}+\boldsymbol{\epsilon}_{\mathbf{x}_{k}}) \big\Vert^2}_{:=\clubsuit}	.\nonumber
	\end{align}
	We calculate each of the above three parts respectively
		\setlength{\jot}{6pt} 
		\begin{align}
		\spadesuit=&-\eta_{k} \big\langle \nabla f(\mathbf{x}_{k}),(1-\lambda_{k}) \mathbf{g}(\mathbf{x}_{k}) \big\rangle \nonumber \\
	=& -\eta_{k}(1-\lambda_{k}) \left [ \big\langle \nabla f(\mathbf{x}_k),\mathbf{g}(\mathbf{x}_{k})-\nabla f(\mathbf{x}_{k}) \big\rangle\right]	-\eta_{k}(1-\lambda_{k})\big\Vert \nabla f(\mathbf{x}_{k}) \big\Vert^2, \label{ineq.2}	\\
		\heartsuit=&-\eta_{k}\big\langle \nabla f(\mathbf{x}_{k}),\lambda_{k} \mathbf{g}(\mathbf{x}_{k}+\boldsymbol{\epsilon}_{\mathbf{x}_{k}}) \big\rangle \nonumber \\
		=& -\eta_{k} \lambda_{k} \big\langle \nabla f(\mathbf{x}_k),\mathbf{g}(\mathbf{x}_{k}+\boldsymbol{\epsilon}_{\mathbf{x}_{k}})-\mathbf{g}(\mathbf{x}_{k})\big\rangle  -\eta_{k} \lambda_{k} \big\langle \nabla f(\mathbf{x}_k),\mathbf{g}(\mathbf{x}_{k}) - \nabla f(\mathbf{x}_k) \big\rangle - \eta_{k} \lambda_{k}\big\Vert \nabla f(\mathbf{x}_{k}) \big\Vert^2 		\nonumber	\\
		\stackrel{(a)}{\leq} & \eta_{k} \lambda_{k}L\rho \| \nabla f(\mathbf{x}_{k}) \| - \eta_{k} \lambda_{k}\| \nabla f(\mathbf{x}_{k}) \|^2 -\eta_{k} \lambda_{k} \big\langle \nabla f(\mathbf{x}_{k}),\mathbf{g}(\mathbf{x}_{k})-\nabla f(\mathbf{x}_{k}) \big\rangle, \label{ineq.3}	\\
		\clubsuit=&\frac{L}{2} \eta^2_k \Vert (1-\lambda_k)\mathbf{g}(\mathbf{x}_k)+\lambda_k \mathbf{g}(\mathbf{x}_k+\boldsymbol{\epsilon}_{\mathbf{x}_{k}})\Vert^2	\nonumber \\
		\stackrel{(b)}{\leq}& L\eta_k^2\left [ (1-\lambda_k)^2\Vert \mathbf{g}(\mathbf{x}_k) \Vert^2+\lambda_k^2 \Vert \mathbf{g}(\mathbf{x}_k+\boldsymbol{\epsilon}_{\mathbf{x}_{k}}) \Vert^2\right ], \label{ineq.4}
		\end{align}
		where the inequality (a) uses Lipschitz continuity $\|\mathbf g(\mathbf x_k+\boldsymbol{\epsilon}_{\mathbf{x}_{k}})-\mathbf g(\mathbf x_k)\|\leq L\| \boldsymbol{\epsilon}_{\mathbf{x}_{k}}\|\ $and the Cauchy's inequality $\langle \mathbf{x},\mathbf{y} \rangle \leq \|\mathbf{x}\| \cdot \| \mathbf{y}\|$.\ The inequality (b) depends on the inequality $\| \mathbf{a}+\mathbf{b}\|^2 \leq 2\|\mathbf a\|^2+2\|\mathbf b\|^2$.
		Taking expectation on \eqref{ineq.2}, \eqref{ineq.3} and \eqref{ineq.4}, we can further get
		\setlength{\jot}{6pt} 
		\begin{align}
			\mathbb{E}[\spadesuit]\stackrel{(c)}{=}&-\eta_k(1-\lambda_k)\mathbb{E} \big [\Vert \nabla f(\mathbf{x}_k) \Vert^2 \big ],  \label{eq.1}    \\
			\mathbb{E}[\heartsuit]\leq& \eta_k\lambda_k L\rho\mathbb{E}\big [ \Vert \nabla f(\mathbf{x}_k) \Vert \big ]- \eta_k \lambda_k \mathbb{E} \big [\Vert \nabla f(\mathbf{x}_k) \Vert^2 \big ] 	\nonumber \\
			\stackrel{(d)}{\leq}&\frac{L}{2}\rho^2+\frac{L}{2}\eta_k^2\lambda_k^2   \big[\mathbb{E}[\Vert\nabla f(\mathbf{x}_k)\Vert] \big]^2-\eta_k\lambda_k\mathbb{E}\big[\Vert \nabla f(\mathbf{x}_k) \Vert^2 \big ]	\nonumber	\\
			\stackrel{(e)}{\leq}&\!\frac{L}{2}\rho^2+\frac{L}{2}\eta_k^2\lambda_k^2 \mathbb{E} \big [\Vert \nabla \!f(\mathbf{x}_k) \Vert^2 \big ]-\eta_k\lambda_k\mathbb{E}\big[\Vert \nabla \!f(\mathbf{x}_k) \Vert^2 \big ], 	\label{ineq.5}\\
			\mathbb{E}[\clubsuit]\leq &L\eta_k^2\left [ (1\!-\!\lambda_k)^2\mathbb{E}[\Vert \mathbf{g}(\mathbf{x}_k) \Vert^2]\!+\!\lambda_k^2 \mathbb{E}[\Vert \mathbf{g}(\mathbf{x}_k+\boldsymbol{\epsilon}_{\mathbf{x}_{k}}) \Vert^2]\right] \nonumber	\\
			\stackrel{(f)}{\leq} & L\eta_k^2\Big [ (1\!-\!\lambda_k)^2\big(\sigma^2+\mathbb{E}[\Vert \nabla f(\mathbf{x}_k)\Vert^2]\big)+\lambda_k^2\big(2L^2\rho^2+2\sigma^2+2\mathbb{E}[\Vert \nabla f(\mathbf{x}_k) \Vert^2]\big)\Big], \label{ineq.6}
			\end{align}
		where the equality (c) utilizes that $\mathbf{g}(\mathbf{x}_k)$ is the unbiased estimation of $\nabla f(\mathbf x_k).$\ The inequality (d) leverages the basic inequality $\frac{a+b}{2} \geq \sqrt{ab} $.\ The inequality (e) comes from the nonnegative variance property $\mathrm{Var}(\mathbf X)=\mathbb{E}[\mathbf X^2 ]-  \big[\mathbb{E}\left [ \mathbf{X} \right ]\big]^2 \geq 0$. A combination of the following results \eqref{exp.1} and \eqref{exp.2} leads to (f).
    After a simple mathematical derivation of $\Vert \mathbf{g}(\mathbf{x}_k)\Vert^2$, the following results could be derived: 
		\setlength{\jot}{6pt} 
		\begin{align}
			\Vert \mathbf{g}(\mathbf{x}_k)\Vert^2 &=\Vert \mathbf{g}(\mathbf{x}_k)-\nabla f(\mathbf{x}_k)+\nabla f(\mathbf{x}_k)\Vert^2    \nonumber \\
			&= \Vert \mathbf{g}(\mathbf{x}_k)-\nabla f(\mathbf{x}_k)\Vert^2+\Vert \nabla f(\mathbf{x}_k)\Vert^2+2\big \langle \mathbf{g}(\mathbf{x}_k)-\nabla f(\mathbf{x}_k),\nabla f(\mathbf{x}_k)\big \rangle.\label{exp}	
		\end{align}
	Taking expectation on \eqref{exp}, then we derive
		\begin{align}
				\mathbb{E}[\Vert\mathbf{g}(\mathbf{x}_k)\Vert^2]&\leq\sigma^2+\mathbb{E} \big [\Vert \nabla f(\mathbf{x}_k) \Vert^2 \big].\label{exp.1}
			\end{align}
        After a plain derivation of $\Vert \mathbf{g}(\mathbf{x}_k+\boldsymbol{\epsilon}_{\mathbf{x}_{k}})\Vert^2$, we have
		\begin{align}
			\Vert \mathbf{g}(\mathbf{x}_k+\boldsymbol{\epsilon}_{\mathbf{x}_{k}})\Vert^2               &=\Vert\mathbf{g}(\mathbf{x}_k+\boldsymbol{\epsilon}_{\mathbf{x}_{k}})-\mathbf{g}(\mathbf{x}_k)+\mathbf{g}(\mathbf{x}_k)\Vert^2 \nonumber \\
			&\leq 2\Vert\mathbf{g}(\mathbf{x}_k+\boldsymbol{\epsilon}_{\mathbf{x}_{k}})-\mathbf{g}(\mathbf{x}_k)\|^2+2\|\mathbf{g}(\mathbf{x}_k)\Vert^2 	\nonumber \\
			&\leq 2L^2\rho^2+2\|\mathbf{g}(\mathbf{x}_k)\Vert^2.		\label{exp2}	
		\end{align}
	Taking expectation on \eqref{exp2} gives us
		\begin{align}
			\mathbb{E}[ \Vert \mathbf g(\mathbf x_k+\boldsymbol{\epsilon}_{\mathbf{x}_{k}})\Vert^2] &\leq 2L^2\rho^2+2\mathbb E[\|\mathbf{g}(\mathbf{x}_k)\Vert^2].	\label{exp.2}
		\end{align}
	Taking expectation on \eqref{ineq.1} and utilizing the results of \eqref{eq.1}, \eqref{ineq.5}, \eqref{ineq.6}, \eqref{exp.1} and \eqref{exp.2}, we can get
	\setlength{\jot}{6pt} 
		\begin{align}\label{ineq.b}
				\mathbb E\left[f(\mathbf{x}_{k+1})-f(\mathbf{x}_k)\right]\leq& \mathbb E[\spadesuit]+\mathbb E[\heartsuit]+\mathbb E[\clubsuit]	\nonumber	\\	
				\leq& -\eta_k(1-\lambda_k)\mathbb{E} \big [\Vert \nabla f(\mathbf{x}_k) \Vert^2 \big ]+\frac{L}{2}\rho^2+\frac{L}{2}\eta_k^2\lambda_k^2 \mathbb{E} \big [\Vert \nabla f(\mathbf{x}_k) \Vert^2 \big ]		-\eta_k\lambda_k\mathbb{E}\big[\Vert \nabla f(\mathbf{x}_k) \Vert^2 \big ]\nonumber \\
				&+L\eta_k^2\Big [ (1-\lambda_k)^2\big(\sigma^2+\mathbb{E}[\Vert \nabla f(\mathbf{x}_k)\Vert^2]\big)+\lambda_k^2\big(2L^2\rho^2+2\sigma^2+2\mathbb{E}[\Vert \nabla f(\mathbf{x}_k) \Vert^2]\big)\Big].
		\end{align}
	Rearranging these terms and dividing the constant $\eta_{k}$ on both sides of \eqref{ineq.b} yields 
			
	\begin{align}
		\underbrace{ \left [ 1-\frac{5}{2}L\eta_{k}\lambda_{k}^2-L\eta_{k}(1-\lambda_{k})^2 \right ]}_{:=M_k} \mathbb E[\|\nabla f(\mathbf x_k)\|^2] \leq&2L\eta_{k}\lambda_{k}^2\sigma^2+\frac{L\rho^2}{2\eta_{k}}+L\eta_{k}(1-\lambda_{k})^2\sigma^2	          \nonumber \\
		&+2L^3\eta_{k}\lambda_{k}^2\rho^2+\frac{\mathbb E[f(\mathbf x_k)-f(\mathbf x_{k+1})]}{\eta_{k}}.  \nonumber
	\end{align}
	It holds that $\lambda_{k}\in(0,1)$ according to the update rule in Algorithm 1. Thus we have $1-\frac{5}{2}L\eta_{k}\leq M_k=1-\frac{5}{2}L\eta_{k}\lambda_{k}^2-L\eta_{k}(1-\lambda_{k})^2\leq 1-\frac{5}{7}L\eta_{k}$, $M_k\geq \nu=1-\frac{5L\eta_0}{2\sqrt K}$ with $\eta_k=\frac{\eta_0}{\sqrt K}$ and $\rho=\frac{\rho_0}{\sqrt K}$. Consequently, we further get
			
	\begin{align}\label{ineq.d}
			\mathbb E[\|\nabla f(\mathbf x_k)\|^2]\leq& \frac{1}{\nu} \Big [ 2L\eta_{k}\sigma^2+\frac{L\rho^2}{2\eta_{k}}+L\eta_{k}\sigma^2+2L^3\eta_{k}\rho^2	+\frac{\mathbb E[f(\mathbf x_k)-f(\mathbf x_{k+1})]}{\eta_{k}} \Big ]		\nonumber\\
			=&\frac{1}{\nu} \Big [ \frac{2L\eta_{0}\sigma^2}{\sqrt K}+\frac{L\rho_0^2}{2\eta_0\sqrt{K}}+\frac{L\eta_0\sigma^2}{\sqrt K}	+\frac{2L^3\eta_0\rho_0^2}{K^\frac{3}{2}}+\frac{\mathbb E[f(\mathbf x_k)-f(\mathbf x_{k+1})]}{\eta_0}\sqrt K \Big ].
	\end{align}
Summing \eqref{ineq.d} from $k=0 \ \text{to}\  K-1$, we have  
	\begin{align}\label{ineq.e}
		\dfrac{1}{K}\sum_{k=0}^{K-1}\mathbb E[\|\nabla f(\mathbf x_k)\|^2]\leq& \frac{1}{\nu} \Big [ \frac{2L\eta_{0}\sigma^2}{\sqrt K}+\frac{L\rho_0^2}{2\eta_0\sqrt{K}}+\frac{L\eta_0\sigma^2}{\sqrt K}	+\frac{2L^3\eta_0\rho_0^2}{K^\frac{3}{2}}+\frac{\mathbb E[f(\mathbf x_0)-f(\mathbf x_K)]}{\eta_0\sqrt K} \Big ] \nonumber	\\
		\leq& \frac{1}{\nu}\Big [ \frac{f(\mathbf x_0)-f_{\mathrm{inf}}}{\eta_0 \sqrt K}+\frac{L\rho_0^2}{2\eta_0\sqrt K}+\frac{3L\eta_0\sigma^2}{\sqrt K}+\frac{2L^3\eta_0\rho_0^2}{K^\frac{3}{2}}\Big ].
	\end{align}
	As to $\frac{1}{K}\displaystyle\sum_{k=0}^{K-1}\mathbb E[\|\nabla f(\mathbf x_k+\boldsymbol{\epsilon}_{\mathbf{x}_{k}})\|^2]$, which could be derived by leveraging on \eqref{ineq.e}. After an ordinary derivation of $\|\nabla f(\mathbf x_k+\boldsymbol{\epsilon}_{\mathbf{x}_{k}})\|^2$, we are then led to
	\begin{align}
		\|\nabla f(\mathbf x_k+\boldsymbol{\epsilon}_{\mathbf{x}_{k}})\|^2&=\|\nabla f(\mathbf x_k+\boldsymbol{\epsilon}_{\mathbf{x}_{k}})-\nabla f(\mathbf x_k)+\nabla f(\mathbf x_k)\|^2	\nonumber	\\
		&\leq 2\|\nabla f(\mathbf x_k+\boldsymbol{\epsilon}_{\mathbf{x}_{k}})-\nabla f(\mathbf x_k)\|^2+2\|\nabla f(\mathbf x_k)\|^2		\nonumber	\\
		&\leq 2L^2\rho^2+2\|\nabla\! f(\mathbf x_k)\|^2. \label{exp.3}
	\end{align}
	Utilizing \eqref{exp.3}, we have
	\setlength{\jot}{6pt} 
		\begin{align}
			\dfrac{1}{K}\sum_{k=0}^{K-1}\mathbb E[\|\nabla f(\mathbf x_k+\boldsymbol{\epsilon}_{\mathbf{x}_{k}})\|^2] &\leq 2L^2\rho^2+\dfrac{2}{K}\sum_{k=0}^{K-1}\mathbb E[\|\nabla f(\mathbf x_k)\|^2]		\nonumber	\\
			&\leq\frac{2}{\nu} \left [\frac{f(\mathbf x_0)-f_{\mathrm{inf}}}{\eta_0\sqrt{K}}+\frac{L\rho_0^2}{2\eta_0\sqrt K}+\frac{3L\eta_0\sigma^2}{\sqrt K}+\frac{2L^3\eta_0\rho_0^2}{K^\frac{3}{2}} \right]+\frac{2L^2\rho_0^2}{K} .
	\end{align}

\subsection{Hyperparameters for experiements}
	The hyperparameters throughout the experiment are noteworthy. Since every model is only trained for 100  epochs, we set a slightly larger initial learning rate to ensure that all models converge after training 100 epochs.
	Referring to the hyperparameters in the relevant literature \cite{DBLP:conf/nips/LiG23, DBLP:conf/iclr/DuYFZZGT22, DBLP:conf/iclr/ForetKMN21, kwon2021asam, li2024friendly, mi2022make}, and combining them with the changes in experimental results we observed while finetuning the hyperparameters, the hyperparameters in experiments are shown in Table \ref{hyperparameters}. Following \cite{DBLP:conf/nips/LiG23}, VaSSO adopts $\theta=0.9$. To reduce the effort of finetuning the hyperparameters, we take $\lambda_0=1, \delta=0.01$ for SAMAR, and weight decay is 0.0005 in all experiments.

\begin{table*}
		\centering
		\caption{Hyperparameters setting used to produce the results of CIFAR-10/CIFAR-100  }
		\label{hyperparameters}
		\begin{tabular}{|c|c|c|c|c|c|c|c|}
			\hline
			Dataset & Model & Optimizer & Lr & $\rho$ & $\theta$ & $\gamma$ & $\chi$ \\
                \hline
                \multirow{8}{*}{CIFAR-10} & \multirow{4}{*}{ResNet-34} & SAMAR & 0.3 & 0.10 & - & 1.550 & 1.100 \\
                
                        &       &   SGD     & 0.3 &   -  & -   & - & - \\
                        &       &   SAM     & 0.3 & 0.10 & -   & - & - \\
                        &       &   VaSSO   & 0.3 & 0.10 & 0.9 & - & - \\
                        \cline{2-8}
                        &       \multirow{4}{*}{Wide-Resnet-34-10} & SAMAR & 0.1 & 0.10 & - & 1.400 & 1.050 \\
                        &       &   SGD     & 0.1 &   -  & -   & - & - \\
                        &       &   SAM     & 0.1 & 0.10 & -   & - & - \\
                        &       &   VaSSO   & 0.1 & 0.10 & 0.9 & - & - \\
                \hline
                \multirow{8}{*}{CIFAR-100} & \multirow{4}{*}{ResNet-34} & SAMAR & 0.3 & 0.10 & - & 1.400 & 1.075 \\
                        &       &   SGD     & 0.3 &   -  & -   & - & - \\
                        &       &   SAM     & 0.3 & 0.10 & -   & - & - \\
                        &       &   VaSSO   & 0.3 & 0.10 & 0.9 & - & - \\
                        \cline{2-8}
                        &       \multirow{4}{*}{Wide-Resnet-34-10} & SAMAR & 0.3 & 0.15 & - & 1.500 & 1.000 \\
                        &       &   SGD     & 0.3 &   -  & -   & - & - \\
                        &       &   SAM     & 0.3 & 0.15 & -   & - & - \\
                        &       &   VaSSO   & 0.3 & 0.15 & 0.9 & - & - \\
                \hline
		\end{tabular}
	\end{table*}
  
  \end{onecolumn}

\end{document}